\documentclass{article}
\usepackage{spconf,amsmath,graphicx}
\usepackage{cite}
\usepackage{amsmath,amssymb,amsfonts}
\usepackage{algorithmic}
\usepackage{graphicx}
\usepackage{textcomp}
\usepackage{xcolor}
\usepackage{hhline}
\usepackage{booktabs} 
\usepackage{multirow}
\usepackage{multicol}
\usepackage{adjustbox}
\usepackage{colortbl}

\newcommand{\myparagraph}[1]{{\vspace{.2em} \noindent \bf #1}}

\newcommand{\egno}{\textit{e}.\textit{g}.}

\definecolor{Gray}{gray}{0.9}

\title{Rethinking Domain Adaptation and Generalization in the Era of CLIP}
\name{$\text{Ruoyu Feng}^1$, $\text{Tao Yu}^1$, $\text{Xin Jin}^2$, $\text{Xiaoyuan Yu}^3$, $\text{Lei Xiao}^3$, $\text{Zhibo Chen}^1$}

\address{1: University of Science and Technology of China \\
2: Eastern Institute of Technology, Ningbo \\
3: Huawei Cloud}
\begin{document}
%\ninept
%
\maketitle
\begin{abstract}
In recent studies on domain adaptation, significant emphasis has been placed on the advancement of learning shared knowledge from a source domain to a target domain.
Recently, the large vision-language pre-trained model (\textit{i.e.}, CLIP) has shown strong ability on zero-shot recognition, and parameter efficient tuning can further improve its performance on specific tasks.
This work demonstrates that a simple domain prior boosts CLIP's zero-shot recognition in a specific domain. Besides, CLIP's adaptation relies less on source domain data due to its diverse pre-training dataset.
Furthermore, we create a benchmark for zero-shot adaptation and pseudo-labeling based self-training with CLIP.
Last but not least, we propose to improve the task generalization ability of CLIP from multiple unlabeled domains, which is a more practical and unique scenario.
We believe our findings motivate a rethinking of domain adaptation benchmarks and the associated role of related algorithms in the era of CLIP.
\end{abstract}
\begin{keywords}
Unsupervised domain adaptation, Domain generalization, Vision-language pre-trained model, Self-training
\end{keywords}

\section{Introduction}
\label{sec:introduction}
% \IEEEPARstart{U}{nsupervised} domain adaptation is a challenging problem that aims to learn common knowledge from the source domain and adapt it to the target domain. 
% This is a challenging problem because the domain distribution discrepancy are complicated and diverse, decreasing the generalization capability of the learned visual representations. 
% In real-world computer vision applications, domain shift often exists between the training data and the testing data. Such domain distribution discrepancy decrease the generalization capability of the learned visual representations, leading to sub-optimal performances on the testing data.
% Unsupervised domain adaptation and domain generalization aims to solve this problem by transferring common knowledge from a labeled source domain to an unlabeled target domain~\cite{ganin2015unsupervised,long2015learning,pan2010survey}.
In the context of real-world computer vision applications~\cite{he2016deep,girshick2015fast,he2017mask,zhao2017pyramid,lin2017feature,liu2021swin ,ding-24-style,li-24-vqa}, the phenomenon of domain shift is a prevalent challenge, where the training data (source domain) and the testing data (target domain) often exhibit significant differences in their distribution. This discrepancy in domain distributions can severely diminish the generalization capability of the visual representations learned by models, leading to sub-optimal performances when applied to the testing data. To tackle this issue, the fields of Unsupervised Domain Adaptation (UDA) and Domain Generalization (DG) have emerged, focusing on mitigating the adverse effects of domain shift. These approaches~\cite{ganin2015unsupervised,long2015learning,pan2010survey} aim to enhance the robustness and adaptability of computer vision models by transferring common knowledge from a well-labeled source domain to an unlabeled or sparsely labeled target domain, thus preserving performance across diverse conditions. 
% Seminal works in this area~\cite{ganin2015unsupervised,long2015learning,pan2010survey} have laid the groundwork for understanding and addressing domain discrepancies, proposing methodologies that enable models to learn domain-invariant features. These strategies not only facilitate the adaptation of models to new, unseen domains without the need for extensive re-labeling but also contribute significantly to the advancement of computer vision systems capable of operating effectively in dynamic real-world scenarios.

% Pre-training has been very effective for deep neural networks across many visual tasks, providing strong initial representations for transferring learning~\cite{he2020momentum,chen2020improved,chen2020simple,caron2020unsupervised,grill2020bootstrap,chen2021exploring}.

\begin{figure}[htbp]
\centering
\includegraphics[scale=0.4]{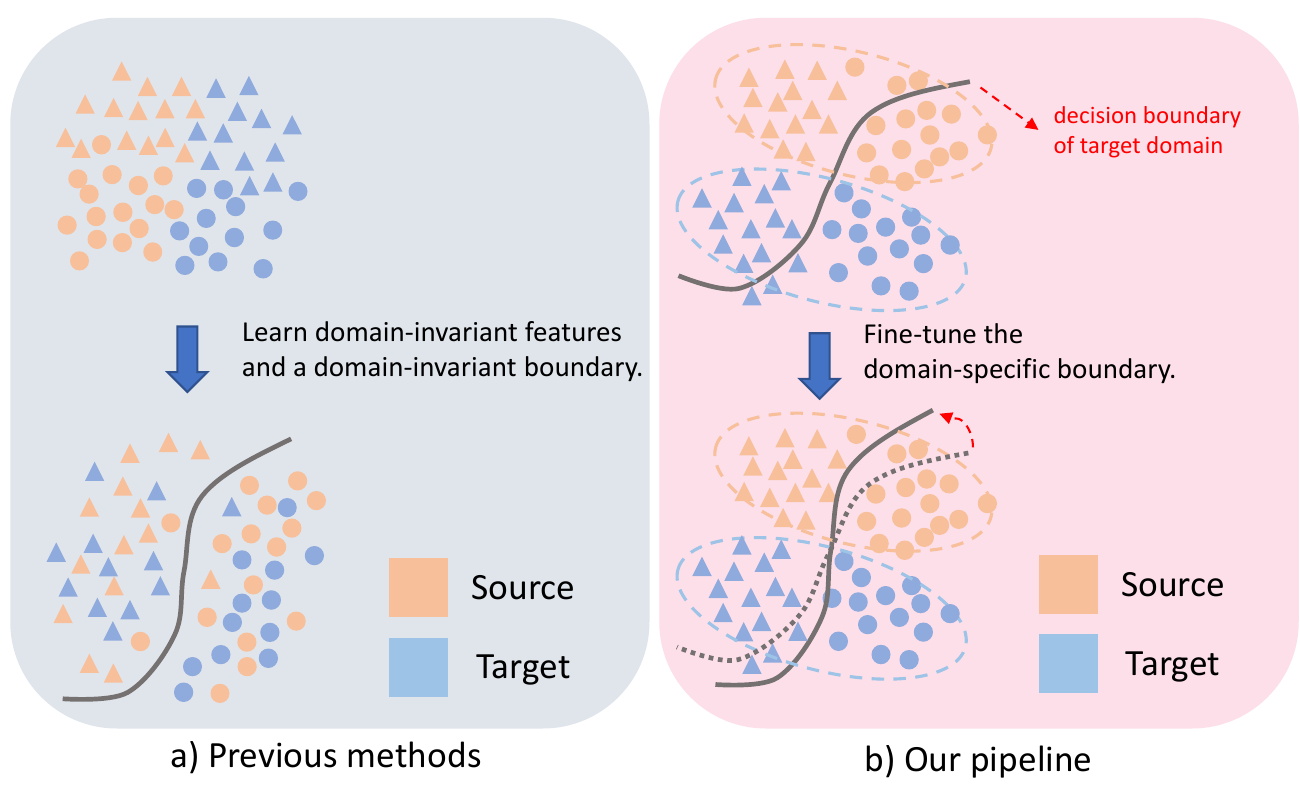}
\caption{Motivation of this paper. Since CLIP has already equipped with strong zero-shot ability, we propose to carefully fine-tune the decision boundary of target domain by learning a task residual with pseudo-labling based self-training.}
\label{Fig.2}
\end{figure}

Recently, vision-language (VL) pre-trained models (\textit{e.g.},  CLIP~\cite{radford2021learning}) has shown promising capability on various computer vision tasks, such as zero-shot learning~\cite{shu2022test}, few-shot learning~\cite{zhou2022learning,zhou2022conditional}, and text-to-image generation~\cite{ramesh2021zero,ramesh2022hierarchical}. 
CLIP consists of two subnetworks, \textit{i.e.}, an image encoder and a text encoder to extract visual representations and text modalities, respectively.
It is trained based on contrastive learning on the 400 million image-text pairs collected from a variety of publicly available sources on the Internet, thus it is born to perform zero-shot classification well on various recognition tasks, even with diverse domain distribution. 
Furthermore, prompt engineering improves CLIP zero-shot performance by customising handcraft prompt templates (\textit{e.g.}, ``\texttt{a photo of a \{class\}}.", ``\texttt{A photo of a \{class\}, a type of pet.}'') to each task. Some works~\cite{zhou2022learning,zhou2022conditional} propose to introduce extra text prompts, treated as learnable parameters, mitigating the potentially suboptimal hand-crafted text prompt templates.

Based on CLIP, DAPL~\cite{ge2022domain} and MPL~\cite{chen2022multi} propose to solve the UDA problem by separating learnable text prompts into domain-agnostic context and domain-specific context. The domain-agnostic context represents general task information and is shared among all images. The domain-specific context represents domain information and is shared in each domain. However, such methods treat one category with different domains as separate categories during the training process, thereby overlooking the inherent relationships that exist between them.
% In addition, the pre-training process of CLIP can be seen as training on a huge and diverse dataset which contains various of source domain data.
% Therefore, we believe that under the traditional UDA setting, CLIP's demand for source domain data is greatly reduced, the improvements are mainly attributed to pseudo-labeling based self-training of target domain data.
% And our experimental results verify the speculation.

In this work, we rethink the unsupervised domain adaptation in the era of CLIP from three perspectives.
First, since CLIP's classifier is encoded representations of text prompts, we observe that the simple domain prior (such as ``infograph'', ``clipart'', and ``quickdraw'') can yield substantial improvements. Furthermore, this enhancement can also prove advantageous for subsequent fine-tuning procedures.
Second, we argue that the pre-training process of CLIP can be seen as training on a huge and diverse dataset that contains various source domain data. Therefore, under the traditional UDA setting, CLIP's demand for source domain data is greatly reduced, the improvements are mainly attributed to pseudo-labeling-based self-training of target domain data. 
Our experimental results also verify this conjecture. 
Furthermore, we build a benchmark for adapting CLIP to a certain task by learning a task residual, which is more efficient than previous full-tuning and prompt-tuning methods. 
Third, we propose a more practical scenario for CLIP-based task generalization with multiple unlabeled source domains. 
And correspondingly, we propose a common-specific knowledge disentangling training strategy to tackle with this problem.
Our experimental results shows that by learning common knowledge from data in different unlabeled domains, the generalization ability of CLIP is significantly enhanced.

The main contributions of this paper are summarised as follows:
\begin{itemize}
\item We observe that simple domain prior can bring significant improvement for CLIP-based zero-shot recognition. This improvement is also helpful for subsequent adaptive fine-tuning.

\item We argue that the success of current CLIP-based UDA methods is attributed to the pseudo-labeling-based self-training and the necessity of source domain data is greatly reduced. Furthermore, we build a benchmark for adapting CLIP to a certain target domain by learning a task residual, which is more efficient than previous prompt-tuning methods.

\item Last but not least, we propose to improve the task generalization ability of CLIP from multiple unlabeled domains, which is more practical and unique for CLIP. 

\end{itemize}

\section{Preliminaries}
\label{sec:preliminary}
In this section, we briefly review CLIP\cite{radford2021learning} and the corresponding zero-shot inference.

\subsection{CLIP Pre-Training}
CLIP consists of an image encoder and a language encoder. 
The image encoder maps the high-dimensional images into a low-dimensional embedding space. The language encoder maps sequences of words into the embedding space with the same dimension of image embedding. 
During the pre-training stage, those two embeddings of images and text sequences are trained to be aligned. 
Given a batch of image-text pairs, the contrastive loss is conducted to maximize the cosine similarity of paired embeddings and minimize the cosine similarity of unpaired embeddings. 
To learn diverse visual concepts and wide range of language expression, CLIP are pre-trained on a huge training dataset consisting of 400 million image-text pairs collected from a variety of publicly available sources on the Internet, which is named as WIT for WebImageText.

\begin{figure*}[htbp]
\centering
\includegraphics[scale=0.4]{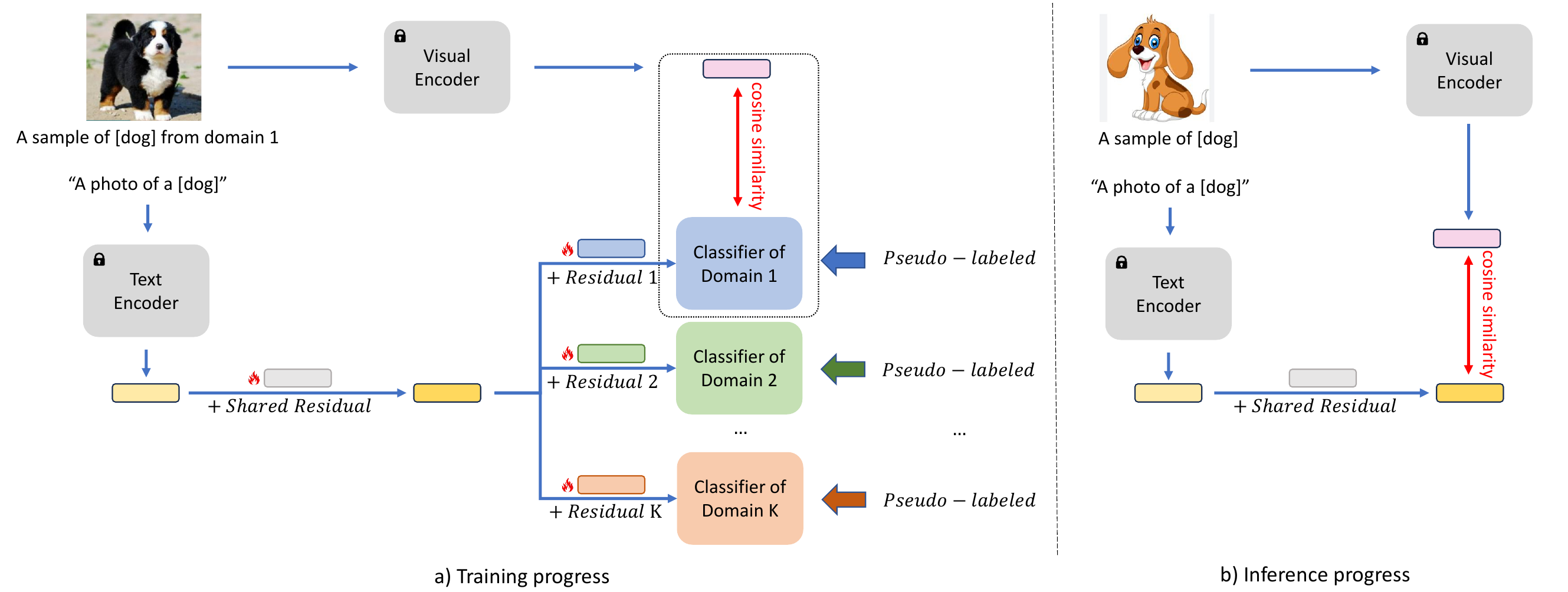}
\caption{\textbf{Training and inference pipelines for learning task information from multiple unlabeled domain data for domain generalization.} a) During training progress, task residual is disentangled into ``shared'' and ``specific''. b) During inference progress, we only take use of share residual which contains common task-adaptive knowledge.}
\label{Fig.2}
\end{figure*}

\subsection{CLIP Zero-Shot Inference}
\label{sec:zero_shot_inference}
Since the pre-training of CLIP is matching images and text descriptions, it fits zero-shot recognition naturally.
More specifically, the zero-shot inference of CLIP is conducted by calculating the similarity between image embedding and text embeddings from the set of possible text descriptions encoded by their respective encoders. 
The cosine similarity of the embeddings is scaled by a temperature parameter $\tau$ and normalized to a probability distribution through a softmax. 
Formally, consider $\boldsymbol{f}$ as the image embedding extracted by the image encoder for an image $\boldsymbol{x}$ and the text embeddings $\{\boldsymbol{t}_{i}\}_{i=1}^{K}$ obtained by feeding text descriptions of $K$ possible classes, \egno, ``\texttt{a photo of a \{class\}}'', into the text encoder. 
The predicted probability is formulated as:
\begin{equation}
\label{eq:naive_clip_prob}
    p(y_i|\boldsymbol{f}) = \frac{\mathrm{exp}(\mathrm{sim}(\boldsymbol{f},\boldsymbol{t}_{i})/\tau)}{\sum_{j=1}^{K}\mathrm{exp}(\mathrm{sim}(\boldsymbol{f},\boldsymbol{t}_{j})/\tau)},
\end{equation}
where $\tau$ denotes the temperature parameter, and $\mathrm{sim}(\cdot,\cdot)$ indicates the cosine similarity function.

\subsection{CLIP-based Prompt Tuning}
% Vision-language (VL) pre-trained models~\cite{radford2021learning,jia2021scaling} use the contrastive loss to align the image embeddings and text embeddings in a common feature space. These vision-language pre-trained models are usually trained on huge and diverse datasets consisted of image-text pairs collected in the website.
% Although VL models show strong performances in zero-shot recognition and great transferability across various intelligent downstream tasks such as few-shot learning~\cite{zhou2022learning,zhou2022conditional}, video retrieval~\cite{luo2021clip4clip}, object detection~\cite{gu2021open}, and semantic segmentation~\cite{zhou2022extract}. 
% In computer vision, prompt tuning is applied to fine-tune the input context tokens of VL models for specific task adaptation. CoOp~\cite{zhou2022learning}, CoCoOp~\cite{zhou2022conditional}, DualCoOp~\cite{sun2022dualcoop}, ProGrad~\cite{xing2022class}, and ProDA~\cite{lu2022prompt} are representative works.
Vision-Language (VL) models have demonstrated impressive capabilities in zero-shot recognition and exhibit significant transferability across a range of intelligent downstream tasks, including few-shot learning~\cite{zhou2022learning,zhou2022conditional}, video retrieval~\cite{luo2021clip4clip}, object detection~\cite{gu2021open}, and semantic segmentation~\cite{zhou2022extract}. In the domain of computer vision, prompt tuning has emerged as a pivotal technique for fine-tuning the input context tokens of VL models to better adapt to specific tasks. Notable examples of this approach include CoOp~\cite{zhou2022learning}, CoCoOp~\cite{zhou2022conditional}, DualCoOp~\cite{sun2022dualcoop}, ProGrad~\cite{xing2022class}, ProDA~\cite{lu2022prompt}, and TaskRes~\cite{yu2022task}, each representing significant contributions to the field. 
The fundamental principle behind these methods involves adapting the pre-trained CLIP model for specific tasks by adding or fine-tuning a minimal number of parameters. This approach allows for efficient customization of the model to new tasks without the need for extensive retraining or overhauling the entire model architecture.

\section{Proposed Method}
\label{sec:approach}

\subsection{Enhance the Zero-Shot Inference with Domain Prior}
\label{sec: domain prior}
% In the zero-shot inference process of CLIP, the standard prompt (``a photo of a \{class\}'') shows robust performances \cite{radford2021learning} on various datasets.
During the zero-shot inference process utilized by CLIP, the conventional prompt format, ``a photo of a {class}'', has been empirically validated to demonstrate robust performance across a variety of datasets, as reported by Radford et al.~\cite{radford2021learning}. This methodology leverages CLIP's extensive training on a diverse corpus of images and text pairs, allowing it to effectively generalize and accurately categorize images into specific classes without requiring direct training on those classes.
% However, this standard prompt cannot provide an equally robust prompt for visual semantic representations of different open-world domains.
While the standard prompt ``a photo of a {class}'' has shown robust performance in CLIP's zero-shot inference across various datasets, it falls short in adequately addressing the diverse and complex needs of different open-world domains. These domains often entail a wide array of visual nuances that a generic prompt might not capture effectively.
In practical scenarios, test data for the target domain can often be summarised with salient descriptions. 
% Obtaining these descriptions does not take much effort, but it can benefit the classification process by providing additional information that language affords to the CLIP text encoder. 
Obtaining these detailed and contextually rich descriptions requires minimal effort yet substantially enriches the classification process by supplying the CLIP text encoder with the nuanced information that only language can provide, thereby significantly enhancing its ability to interpret and classify visual content with greater accuracy and relevance.
Specifically, the prompt evolves into ``\texttt{a \{domain description\} photo of a \{class\}}'' instead of naive ``\texttt{a photo of a \{class\}}''.
Formally, consider the text embeddings $\{\boldsymbol{d}_{i}\}_{i=1}^{K}$ obtained by feeding text prompt with domain descriptions of possible classes into the text encoder. The predicted probability mentioned in Equation \ref{eq:naive_clip_prob} changes into:
\begin{equation}
\label{eq:naive_clip_prob}
    p(y_i|\boldsymbol{f}) = \frac{\mathrm{exp}(\mathrm{sim}(\boldsymbol{f},\boldsymbol{d}_{i})/\tau)}{\sum_{j=1}^{K}\mathrm{exp}(\mathrm{sim}(\boldsymbol{f},\boldsymbol{d}_{j})/\tau)},
\end{equation}
where $\tau$ denotes the temperature parameter, and $\mathrm{sim}(\cdot,\cdot)$ indicates the cosine similarity function.

\subsection{Pseudo-Labeling based Self-Training}
\label{Sec: domain prior}
% CLIP is pre-trained on a huge dataset that containing 400 milion image-text pairs, thus 
% Thus, 
% The datasets for pre-training of CLIP is huge and contains 400 milion image-text pairs collected from a variaty of publicly availiable sources on the Internet, which means it can be regarded as a collection of a large amount of datasets from different source domains. 
The datasets used for the pre-training of CLIP are vast, encompassing 400 million image-text pairs sourced from a wide array of publicly available content on the Internet. This extensive collection effectively represents a compilation of numerous datasets from diverse source domains, providing CLIP with a broad and rich foundation for learning. This variety ensures that CLIP is exposed to a wide spectrum of visual and linguistic information, allowing it to develop a robust understanding of the complex relationships between images and their associated textual descriptions. Consequently, this extensive pre-training enables CLIP to achieve remarkable versatility and generalization capabilities across a multitude of tasks and domains.
Therefore, those visual features of CLIP after pretraining are born to be general, with no need to add further domain-invariant constrains mentioned in previous works \cite{long2015learning}. 
% Meanwhile, the importance of source domain data is greatly reduced.
% A simple yet efficient way to adapt CLIP to current task is pseudo-labelling-based self-training. 
The need for specific source domain data is notably reduced thanks to CLIP's extensive pre-training. Adapting CLIP to a new task can be efficiently achieved through pseudo-labeling-based self-training, a method that enhances its applicability with minimal complexity.
More specifically, consider the unlabeled data in target domain, the pseudo label $y^u$ is generated by the maximum predicted probability of the training data $\boldsymbol{x}^u$:
\begin{equation}
    y^u = \mathop{\arg\max}\limits_{k}P(\hat{y}^u=k|\boldsymbol{x}^u), k=\{1,2,...,K\},
\end{equation}
where $y^u$ are chosen from $K$ classes.

% A fixed threshold $\tau$ is used to filter out samples that are not confident enough. 
A predetermined threshold, denoted as $\gamma$, is applied to selectively filter out samples for which the model's confidence does not meet this specified level, ensuring only high-confidence predictions are utilized.
The zero-shot inference mentioned in \ref{sec:zero_shot_inference} is utilized to generate pseudo-labels.
The overall loss function of pseudo-labeling based self-training is written as:
\begin{equation}
 \mathcal{L}_{u} = -\frac{1}{N_u} \sum_{i = 1}^{N_u}\mathbb{I}\{P(\hat y^u_i = y^{u}_i| \boldsymbol{x}^u_i) \ge \gamma \} \log P(\hat y^u_i = y^{u}_i| \boldsymbol{x}^u_i),
\end{equation}
where $\mathbb{I}\{\cdot\}$ is an indicator function, and ${N_u}$ indicates the number of unlabeled data.

\subsection{Task Residual for Tuning Vision-Language Models}
% Parameter efficient tuning \cite{jia2022visual, yu2022task} has shown great potential on fine-tuning VL models for downstream tasks. 
Parameter-efficient tuning, as explored in studies by Jia et al.~\cite{jia2022visual} and Yu et al.~\cite{yu2022task}, has demonstrated significant promise in fine-tuning Vision-Language (VL) models for downstream tasks. These approaches optimize the adaptation of VL models with minimal modifications to their parameters, enhancing their performance on specific tasks while maintaining the integrity of the pre-trained models.
% In this paper, we apply TaskRes\cite{yu2022task} to build the benchmark for its simplicity and strongness.
In this paper, we utilize TaskRes~\cite{yu2022task} as the foundation for our benchmark, selected for its simplicity and robust performance. This approach allows us to systematically evaluate and demonstrate the effectiveness of TaskRes in enhancing Vision-Language model capabilities for specific applications, showcasing its practicality and strength in adapting to diverse tasks.
% Specifically, TaskRes learns task-specific knowledge without being restricted by prior knowledge, which is instantiated by a residual added to the text embedding. 
Specifically, TaskRes is designed to learn task-specific knowledge without being hindered by prior knowledge. This is achieved by introducing a residual component that is added to the text embedding. This innovative approach enables the model to adapt and enhance its performance on specific tasks by incorporating additional, task-relevant information directly into the text representations, thereby enriching the model's understanding and responsiveness to the nuances of the task at hand.
Formally, consider the text embeddings $\{\boldsymbol{t}_{i}\}_{i=1}^{K}$ obtained by feeding text descriptions of $K$ possible classes, then the modified text $\boldsymbol{t}_i'$ embedding of class $i$ with task residual $r_{i}$ can be written as
\begin{equation}
    {\boldsymbol{t}_i}'= \boldsymbol{t}_i + \boldsymbol{r}_i,
\end{equation}
where $\boldsymbol{r}_i \in \mathbb{R}^D$, $D$ indicates the text embedding's dimension.

\subsection{Label-Free Multi-source Domain Generalization}
% In this section, we describe a more practical and challenging scenario that learning task information from multiple unlabeled domains, \textit{i.e.}, label-free multi-source domain generalization. 
In this section, we explore learning from multiple unlabeled domains without labels, a scenario known as label-free multi-source domain generalization. This approach challenges models to generalize across diverse domains, relying on unlabeled data to gain task-relevant insights.
% In real-world applications, it is often the case that a large amount of unlabeled data is collected from multiple different source domains. 
In real-world applications, there frequently exists a vast amount of unlabeled data gathered from a variety of source domains.
% In order to efficiently leverage these data to improve the generalization of the CLIP model for specific tasks, we propose a scheme of decoupling the task residual into domain-shared residual and domain-specific residual during self-training progress, as illustrated in Figure \ref{Fig.2}. 
To effectively utilize this wealth of unlabeled data to enhance the CLIP model's generalization capabilities for specific tasks, we propose a strategy that involves decoupling the task residual into two components during the self-training phase: a domain-shared residual and a domain-specific residual. This approach is graphically represented in Figure \ref{Fig.2}. By separating the residuals, we aim to better capture the commonalities across all domains through the domain-shared residual, while also tailoring the model to the unique characteristics of each domain with the domain-specific residual. This dual-residual scheme is designed to optimize the model's learning and adaptation process, ensuring a more nuanced understanding and application of knowledge across various domains.
% During inference progress, only domain-shared residual is used since it contains common knowledge of current task.
During the inference phase, only the domain-shared residual is utilized, as it encapsulates the common knowledge pertinent to the current task. This approach ensures that the model leverages the generalized insights that are applicable across multiple domains, thus optimizing its performance on the task at hand without being confounded by the specifics of any single domain.
Formally, during training, with a sample of class $i$ from domain $n$, the corresponding text embedding ${\boldsymbol{t}_i}''$ is written as
\begin{equation}
    {{\boldsymbol{t}_i}''= \boldsymbol{t}_i + \boldsymbol{r}^{sh}_i + \boldsymbol{r}^{sp}_i},
\end{equation}
where $\boldsymbol{r}^{sh}_i$ and $\boldsymbol{r}^{sp}_i$ indicate the domain-share residual and domain-specific residual of class $i$, respectively.

\begin{table*}[htbp]
\footnotesize
\centering
\setlength{\tabcolsep}{4pt} 
\caption{Accuracy (\%) on DomainNet\cite{peng2019moment} for unsupervised domain adaptation. The best accuracy is indicated in bold.}
\begin{adjustbox}{max width = 1.0\textwidth}
\begin{tabular}{lccc|ccccccc}
\toprule
\textbf{Methods} & \textbf{Backbone} & \textbf{Pre-train. Data} & \textbf{Source} & \textbf{Clipart} & \textbf{Infograph} & \textbf{Painting} & \textbf{Quickdraw} & \textbf{Sketch} & \textbf{Average} \\
\midrule
CDTrans\cite{xu2021cdtrans} & DeiT-B/16 & ImageNet-1K & Real & 66.2 & 31.0 & 61.5 & 16.2 & 52.9 & 45.6 \\
Broad\cite{kim2022broad} & ViT-B/16 & ALBEF\cite{li2021align} & Real & 73.6 & 37.3 & 65.3 & 12.8 & 62.2 & 50.2 \\
% \midrule
\rowcolor{Gray}
CLIP (zero-shot) & ViT-B/16 & WebImageText\cite{radford2021learning}
 & - & 71.4 & 48.7 & 59.3 & 13.1 & 64.2 & 51.3 \\
\rowcolor{Gray}
CLIP (domain prior) & ViT-B/16 & WebImageText\cite{radford2021learning} & - & 73.8 & 52.6 & 61.3 & \textbf{17.7} & 65.8 & 54.3  \\
\rowcolor{Gray}
CLIP (domain prior \& self-training) & ViT-B/16 & WebImageText\cite{radford2021learning} & - & \textbf{76.5} & \textbf{54.6} & \textbf{66.9} & 17.6 & \textbf{67.5} & \textbf{56.6} \\
\bottomrule
\end{tabular}
\end{adjustbox}
\label{Comparison-DA-DomainNet}
\end{table*}

\begin{table*}[htbp]
\footnotesize
\centering
\caption{Accuracy (\%) on OfficeHome\cite{venkateswara2017deep} for unsupervised domain adaptation. The best accuracy is indicated in bold.}
\begin{adjustbox}{max width = 1.0\textwidth}
\begin{tabular}{lccc|ccccccc}
\toprule
\textbf{Methods} & \textbf{Backbone} & \textbf{Pre-train. Data} & \textbf{Source} & \textbf{Art} & \textbf{Clipart} & \textbf{Product} & \textbf{Average} \\
\midrule
CDTrans\cite{xu2021cdtrans} & DeiT-B/16 & ImageNet-1K & Real & 82.0 & 66.0 & 90.6 & 79.5 \\
Broad\cite{kim2022broad} & ViT-B/16 & ALBEF\cite{li2021align} & Real & 81.7  & 72.5 & 87.2 & 80.5  \\
TVT\cite{yang2023tvt} & ViT-B/16 & ImageNet-1K & Real & 79.1 & 67.2 & 88 & 78.1  \\
TVT\cite{yang2023tvt} & ViT-B/16 & ImageNet-21K & Real & \textbf{85.5}
& \textbf{74.6} & 90.6 & 83.6 \\
\rowcolor{Gray}
CLIP (zero-shot) & ViT-B/16 & WebImageText\cite{radford2021learning}
 & - & 82.7  & 68.7  & 89.9 & 80.4 
 \\
\rowcolor{Gray}
CLIP (domain prior) & ViT-B/16 & WebImageText\cite{radford2021learning} & - & 82.6  & 68.7 & 91.4 & 80.9 \\
\rowcolor{Gray}
CLIP (domain prior \& self-training) & ViT-B/16 & WebImageText\cite{radford2021learning} & - & 84.7 & 74.1 & \textbf{93.4} & \textbf{84.1} \\
\bottomrule
\end{tabular}
\end{adjustbox}
\label{Comparison-DA-OfficeHome}
\end{table*}

\section{Experiments}
We conduct extensive experiments on domain adaptation benchmarks to verify the effectiveness of our proposed method. We next present the datasets used in our experiments, comparisons with baseline methods, ablation studies of our method.
\subsection{Datasets and Experimental Settings}
% We conduct comprehensive experiments on \textbf{DomainNet}\cite{peng2019moment} and \textbf{OfficeHome}\cite{venkateswara2017deep}. 
We carry out extensive experiments on two benchmark datasets: \textbf{DomainNet}~\cite{peng2019moment} and \textbf{OfficeHome}~\cite{venkateswara2017deep}. These datasets are well-regarded in the domain adaptation and generalization research community for their diversity in domain representations and the challenges they present in terms of domain shift and generalization. By testing our approach on these datasets, we aim to thoroughly evaluate the effectiveness of our proposed method in enhancing the generalization capabilities of the CLIP model across different source domains.
The input image size in our experiments is $224\times224$. We use the Adam\cite{kingma2014adam} algorithm to optimize the training process. The learning rate
is set to 3e-4. The batch size is set to 64. We freeze the image encoder and the text encoder of CLIP and only optimize the task residual for 5 epochs and 10 epochs on DomainNet and OfficeHome, respectively.

\subsection{Comparison with Other Methods}
\myparagraph{Domain Adaptation. } 
% We compare our methods with CDTrans\cite{xu2021cdtrans}, TVT\cite{yang2023tvt}, and Broad\cite{kim2022broad}. The results of DomainNet and OfficeHome are presented in Table \ref{Comparison-DA-DomainNet} and Table \ref{Comparison-DA-OfficeHome}.
We benchmark our proposed methods against existing approaches, including CDTrans~\cite{xu2021cdtrans}, TVT~\cite{yang2023tvt}, and Broad~\cite{kim2022broad}. The performance comparisons on DomainNet and OfficeHome datasets are summarized in Table \ref{Comparison-DA-DomainNet} and Table \ref{Comparison-DA-OfficeHome}, respectively. These tables provide a comprehensive overview of how our method stacks up against state-of-the-art approaches in terms of domain adaptation performance across different evaluation metrics.
It can be observed that the ``domain prior'' described in Section \ref{sec: domain prior} brings significant gains (3.0\% on DomainNet and 0.5\% on OfficeHome on average). Self-training described in Section \ref{Sec: domain prior} can further improve the accuracy obviously. 
% On the other side, all results based on CLIP surpass the SOTA methods CDTrans\cite{xu2021cdtrans}, TVT\cite{yang2023tvt} and Broad\cite{kim2022broad}, without any labeled data on both source and target domain. 
On the flip side, all results derived from CLIP outperform state-of-the-art methods, including CDTrans~\cite{xu2021cdtrans}, TVT~\cite{yang2023tvt}, and Broad~\cite{kim2022broad}, without the need for any labeled data in either the source or target domains. This underscores the efficacy and versatility of CLIP in domain adaptation tasks, where it demonstrates superior performance even in the absence of labeled data, setting a new benchmark for unsupervised domain adaptation methodologies.
% We owe the performance improvement to the strong capability of CLIP and efficient tuning potential of task residual. 
% And furthermore, the effectiveness of our approach further confirms the importance of rethinking the problem of Domain Adaptation in the era of CLIP.
The observed performance improvements can be attributed to the robust capabilities of CLIP and the efficient tuning potential of the task residual component. Moreover, our approach's effectiveness serves as further validation of the significance of reevaluating the domain adaptation problem in light of the emergence of CLIP. This underscores the transformative impact of leveraging pre-trained vision-language models in addressing domain adaptation challenges, paving the way for novel methodologies and approaches that capitalize on the unique strengths of these models in adapting to diverse and complex real-world scenarios.

\begin{table}[t]
\footnotesize
\centering
\caption{Accuracy (\%) for label-free multi-source domain generalization. The best accuracy is indicated in bold.}
\begin{adjustbox}{max width = 1.0\textwidth}
\begin{tabular}{lcc|cc}
\toprule
% \textbf{Methods} & \textbf{Backbone} & \textbf{Pre-train. Data} & \textbf{DomainNet} & \textbf{OfficeHome}\\
\textbf{Methods} & \textbf{Backbone} & \textbf{Pre-train. Data} & \textbf{Do.} & \textbf{Of.}\\
\midrule
DoPrompt\cite{zheng2022prompt} & ViT-B/16 & ImageNet-1K & 48.3 & 83.2 \\
\rowcolor{Gray}
CLIP (zero-shot) & ViT-B/16 & WebImageText\cite{radford2021learning}
 & 53.1 & 86.7 \\
\rowcolor{Gray}
CLIP (ours) & ViT-B/16 & WebImageText\cite{radford2021learning} & \textbf{55.0} & \textbf{87.2} \\
\bottomrule
\end{tabular}
\end{adjustbox}
\label{Comparison: label-free multi-source domain generalization}
\end{table}

\begin{table}[t]
\footnotesize
\centering
\caption{Ablation study on parameter-efficient tuning methods for unsupervised domain adaptation. The best accuracy is indicated in bold.}
\begin{adjustbox}{max width = 1.0\textwidth}
\begin{tabular}{c|ccccccc}
\toprule
\multicolumn{1}{l|}{\textbf{Methods}} & \textbf{clp} & \textbf{inf} & \textbf{pnt} & \textbf{qdr} & \textbf{rel} & \textbf{skt} & \textbf{Avg.} \\ 
% \hline
\midrule
Zero-shot                             & 67.8        & 42.6        & 55.2        & \textbf{12.5}        & 81.7         & 58.6         & 53.0         \\
Full-Tuning                                 & 52.2        & 22.5        & 45.3        & 6.0         & 83.9         & 34.7        & 40.8        \\
\multicolumn{1}{l|}{Adapter\cite{gao2021clip}}          & 65.7         & 37.6        & 58.7        & 12.0        & 87.4        & 53.9        & 52.6         \\
VPT\cite{jia2022visual}                                   & 63.5         & 37.5        & 57.2        & 9.5         & \textbf{87.5}        & 52.0        & 51.2         \\
\rowcolor{Gray}
TaskRes\cite{yu2022task}                               & \textbf{71.8}        & \textbf{46.3}        & \textbf{63.7}        & 12.0        & 82.6        & \textbf{62.0}        & \textbf{56.4}    \\
\bottomrule
\end{tabular}
\end{adjustbox}
\label{Ablation: parameter efficient tuning methods}
\end{table}

\begin{table}[t]
\footnotesize
\centering
\caption{Ablation study on the type of residual for label-free multi-source domain generalization. The best accuracy is indicated in bold.}
\begin{tabular}{c|ccccccc}
\toprule
\textbf{Methods}  & \textbf{clp} & \textbf{inf} & \textbf{pnt} & \textbf{qdr} & \textbf{rel} & \textbf{skt} & \textbf{Avg.} \\ 
\midrule
Zero-shot & 67.8        & \textbf{42.6}        & 55.2         & 12.5         & \textbf{81.7}        & 58.6        & 53.0      \\
Com. & 67.9        & 39.8        & 61.3        & 12.1        & 81.2        & 59.4        & 53.6     \\
\rowcolor{Gray}
Ours             & \textbf{68.6}        & 40.8        & \textbf{63.0}        & \textbf{16.1}        & 81.3        & \textbf{60.4}        & \textbf{55.0} \\
\bottomrule
\end{tabular}
\label{Ablation: residual type for label-free multi-source domain generalization }
\end{table}

\subsection{Ablation Study}
\myparagraph{Parameter-Efficient Tuning Methods.} 
% We compare the results of different parameter-efficient tuning methods based on CLIP\cite{radford2021learning}, which are shown in Table \ref{Ablation: parameter efficient tuning methods}.
We conduct a comparative analysis of the results obtained from various parameter-efficient tuning methods based on CLIP~\cite{radford2021learning}, as summarized in Table \ref{Ablation: parameter efficient tuning methods}. This comparison allows us to assess the relative efficacy and performance of different tuning strategies in fine-tuning CLIP for specific tasks.
% It can be observed that the TaskRes\cite{yu2022task} is more suitable for the scenario of domain adaptation.
The results indicate that TaskRes~\cite{yu2022task} is particularly well-suited for domain adaptation scenarios.

\myparagraph{Residual Type for Label-Free Multi-Source Domain Generalization.} 
% We evaluate the effectiveness of our proposed disentangling the task residual into domain-shared residual and domain-specific residual on DomainNet dataset. 
We assess the effectiveness of our proposed method, which involves disentangling the task residual into domain-shared residual and domain-specific residual, on the DomainNet dataset.
% As shown in Table \ref{Ablation: residual type for label-free multi-source domain generalization }, our proposed method achieves remarkable improvements compared to only using a common task residual (Com.) in this scenario. 
As demonstrated in Table \ref{Ablation: residual type for label-free multi-source domain generalization }, our proposed method exhibits remarkable improvements compared to solely utilizing a common task residual (Com.) in this label-free multi-source domain generalization scenario.

\section{Conclusion}
% In this paper, we propose to rethink the problem of unsupervised domain adaptation on the era of CLIP. Equipped with strong capability of CLIP and efficient tuning methods, our proposed domain prior and simple self-training can achieves much higher performances compared to traditional UDA methods even without any source labeled data. Moreover, we introduce a more practical scenario of label-free multi-source domain generalization and correspondingly design the method of tuning domain-share residual and domain-specific residual. Comprehensive experimental results shows the effectiveness of our observation and proposed method.
In this paper, we propose a novel perspective on the problem of unsupervised domain adaptation in the era of CLIP. Leveraging the powerful capabilities of CLIP and efficient tuning methods, our approach, which includes domain prior and simple self-training techniques, achieves significantly improved performance compared to traditional UDA methods, even in the absence of labeled source data. Additionally, we introduce a more practical scenario of label-free multi-source domain generalization and develop a corresponding method for tuning domain-shared and domain-specific residuals. Our comprehensive experimental results validate the effectiveness of our observations and proposed methodology.

% \subsection{{Acknowledgement.}
\myparagraph{Acknowledgement.}
This work was supported in part by NSFC under Grant 62371434, 62021001.This work was supported in part by NSFC under Grant 62371434, 62021001.

\bibliographystyle{IEEEbib}
\bibliography{strings,refs}

\begin{thebibliography}{10}

\bibitem{he2016deep}
Kaiming He, Xiangyu Zhang, Shaoqing Ren, and Jian Sun,
\newblock ``Deep residual learning for image recognition,''
\newblock in {\em Proceedings of the IEEE conference on computer vision and pattern recognition}, 2016, pp. 770--778.

\bibitem{girshick2015fast}
Ross Girshick,
\newblock ``Fast r-cnn,''
\newblock in {\em Proceedings of the IEEE international conference on computer vision}, 2015, pp. 1440--1448.

\bibitem{he2017mask}
Kaiming He, Georgia Gkioxari, Piotr Doll{\'a}r, and Ross Girshick,
\newblock ``Mask r-cnn,''
\newblock in {\em Proceedings of the IEEE international conference on computer vision}, 2017, pp. 2961--2969.

\bibitem{zhao2017pyramid}
Hengshuang Zhao, Jianping Shi, Xiaojuan Qi, Xiaogang Wang, and Jiaya Jia,
\newblock ``Pyramid scene parsing network,''
\newblock in {\em Proceedings of the IEEE conference on computer vision and pattern recognition}, 2017, pp. 2881--2890.

\bibitem{lin2017feature}
Tsung-Yi Lin, Piotr Doll{\'a}r, Ross Girshick, Kaiming He, Bharath Hariharan, and Serge Belongie,
\newblock ``Feature pyramid networks for object detection,''
\newblock in {\em Proceedings of the IEEE conference on computer vision and pattern recognition}, 2017, pp. 2117--2125.

\bibitem{liu2021swin}
Ze~Liu, Yutong Lin, Yue Cao, Han Hu, Yixuan Wei, Zheng Zhang, Stephen Lin, and Baining Guo,
\newblock ``Swin transformer: Hierarchical vision transformer using shifted windows,''
\newblock in {\em Proceedings of the IEEE/CVF international conference on computer vision}, 2021, pp. 10012--10022.

\bibitem{ding-24-style}
Zhicheng Ding, Panfeng Li, Qikai Yang, Siyang Li, and Qingtian Gong,
\newblock ``Regional style and color transfer,''
\newblock {\em arXiv preprint arXiv:2404.13880}, 2024.

\bibitem{li-24-vqa}
Panfeng Li, Qikai Yang, Xieming Geng, Wenjing Zhou, Zhicheng Ding, and Yi~Nian,
\newblock ``Exploring diverse methods in visual question answering,''
\newblock {\em arXiv preprint arXiv:2404.13565}, 2024.

\bibitem{ganin2015unsupervised}
Yaroslav Ganin and Victor Lempitsky,
\newblock ``Unsupervised domain adaptation by backpropagation,''
\newblock in {\em ICML}. PMLR, 2015, pp. 1180--1189.

\bibitem{long2015learning}
Mingsheng Long, Yue Cao, Jianmin Wang, and Michael Jordan,
\newblock ``Learning transferable features with deep adaptation networks,''
\newblock in {\em ICML}. PMLR, 2015, pp. 97--105.

\bibitem{pan2010survey}
Sinno~Jialin Pan and Qiang Yang,
\newblock ``A survey on transfer learning,''
\newblock {\em IEEE Transactions on knowledge and data engineering}, vol. 22, no. 10, pp. 1345--1359, 2010.

\bibitem{radford2021learning}
Alec Radford, Jong~Wook Kim, Chris Hallacy, Aditya Ramesh, Gabriel Goh, Sandhini Agarwal, Girish Sastry, Amanda Askell, Pamela Mishkin, Jack Clark, et~al.,
\newblock ``Learning transferable visual models from natural language supervision,''
\newblock in {\em ICML}. PMLR, 2021, pp. 8748--8763.

\bibitem{shu2022test}
Manli Shu, Weili Nie, De-An Huang, Zhiding Yu, Tom Goldstein, Anima Anandkumar, and Chaowei Xiao,
\newblock ``Test-time prompt tuning for zero-shot generalization in vision-language models,''
\newblock {\em arXiv preprint arXiv:2209.07511}, 2022.

\bibitem{zhou2022learning}
Kaiyang Zhou, Jingkang Yang, Chen~Change Loy, and Ziwei Liu,
\newblock ``Learning to prompt for vision-language models,''
\newblock {\em IJCV}, vol. 130, no. 9, pp. 2337--2348, 2022.

\bibitem{zhou2022conditional}
Kaiyang Zhou, Jingkang Yang, Chen~Change Loy, and Ziwei Liu,
\newblock ``Conditional prompt learning for vision-language models,''
\newblock in {\em CVPR}, 2022, pp. 16816--16825.

\bibitem{ramesh2021zero}
Aditya Ramesh, Mikhail Pavlov, Gabriel Goh, Scott Gray, Chelsea Voss, Alec Radford, Mark Chen, and Ilya Sutskever,
\newblock ``Zero-shot text-to-image generation,''
\newblock in {\em ICML}. PMLR, 2021, pp. 8821--8831.

\bibitem{ramesh2022hierarchical}
Aditya Ramesh, Prafulla Dhariwal, Alex Nichol, Casey Chu, and Mark Chen,
\newblock ``Hierarchical text-conditional image generation with clip latents,''
\newblock {\em arXiv preprint arXiv:2204.06125}, 2022.

\bibitem{ge2022domain}
Chunjiang Ge, Rui Huang, Mixue Xie, Zihang Lai, Shiji Song, Shuang Li, and Gao Huang,
\newblock ``Domain adaptation via prompt learning,''
\newblock {\em arXiv preprint arXiv:2202.06687}, 2022.

\bibitem{chen2022multi}
Haoran Chen, Zuxuan Wu, and Yu-Gang Jiang,
\newblock ``Multi-prompt alignment for multi-source unsupervised domain adaptation,''
\newblock {\em arXiv preprint arXiv:2209.15210}, 2022.

\bibitem{luo2021clip4clip}
Huaishao Luo, Lei Ji, Ming Zhong, Yang Chen, Wen Lei, Nan Duan, and Tianrui Li,
\newblock ``Clip4clip: An empirical study of clip for end to end video clip retrieval,''
\newblock {\em arXiv preprint arXiv:2104.08860}, 2021.

\bibitem{gu2021open}
Xiuye Gu, Tsung-Yi Lin, Weicheng Kuo, and Yin Cui,
\newblock ``Open-vocabulary object detection via vision and language knowledge distillation,''
\newblock {\em ICLR}, 2021.

\bibitem{zhou2022extract}
Chong Zhou, Chen~Change Loy, and Bo~Dai,
\newblock ``Extract free dense labels from clip,''
\newblock in {\em ECCV}, 2022, pp. 696--712.

\bibitem{sun2022dualcoop}
Ximeng Sun, Ping Hu, and Kate Saenko,
\newblock ``Dualcoop: Fast adaptation to multi-label recognition with limited annotations,''
\newblock {\em arXiv preprint arXiv:2206.09541}, 2022.

\bibitem{xing2022class}
Yinghui Xing, Qirui Wu, De~Cheng, Shizhou Zhang, Guoqiang Liang, and Yanning Zhang,
\newblock ``Class-aware visual prompt tuning for vision-language pre-trained model,''
\newblock {\em arXiv preprint arXiv:2208.08340}, 2022.

\bibitem{lu2022prompt}
Yuning Lu, Jianzhuang Liu, Yonggang Zhang, Yajing Liu, and Xinmei Tian,
\newblock ``Prompt distribution learning,''
\newblock in {\em Proceedings of the IEEE/CVF Conference on Computer Vision and Pattern Recognition}, 2022, pp. 5206--5215.

\bibitem{yu2022task}
Tao Yu, Zhihe Lu, Xin Jin, Zhibo Chen, and Xinchao Wang,
\newblock ``Task residual for tuning vision-language models,''
\newblock {\em arXiv preprint arXiv:2211.10277}, 2022.

\bibitem{jia2022visual}
Menglin Jia, Luming Tang, Bor-Chun Chen, Claire Cardie, Serge Belongie, Bharath Hariharan, and Ser-Nam Lim,
\newblock ``Visual prompt tuning,''
\newblock in {\em Computer Vision--ECCV 2022: 17th European Conference, Tel Aviv, Israel, October 23--27, 2022, Proceedings, Part XXXIII}. Springer, 2022, pp. 709--727.

\bibitem{peng2019moment}
Xingchao Peng, Qinxun Bai, Xide Xia, Zijun Huang, Kate Saenko, and Bo~Wang,
\newblock ``Moment matching for multi-source domain adaptation,''
\newblock in {\em Proceedings of the IEEE/CVF international conference on computer vision}, 2019, pp. 1406--1415.

\bibitem{xu2021cdtrans}
Tongkun Xu, Weihua Chen, Pichao Wang, Fan Wang, Hao Li, and Rong Jin,
\newblock ``Cdtrans: Cross-domain transformer for unsupervised domain adaptation,''
\newblock {\em arXiv preprint arXiv:2109.06165}, 2021.

\bibitem{kim2022broad}
Donghyun Kim, Kaihong Wang, Stan Sclaroff, and Kate Saenko,
\newblock ``A broad study of pre-training for domain generalization and adaptation,''
\newblock {\em arXiv preprint arXiv:2203.11819}, 2022.

\bibitem{li2021align}
Junnan Li, Ramprasaath Selvaraju, Akhilesh Gotmare, Shafiq Joty, Caiming Xiong, and Steven Chu~Hong Hoi,
\newblock ``Align before fuse: Vision and language representation learning with momentum distillation,''
\newblock {\em Advances in neural information processing systems}, vol. 34, pp. 9694--9705, 2021.

\bibitem{venkateswara2017deep}
Hemanth Venkateswara, Jose Eusebio, Shayok Chakraborty, and Sethuraman Panchanathan,
\newblock ``Deep hashing network for unsupervised domain adaptation,''
\newblock in {\em Proceedings of the IEEE conference on computer vision and pattern recognition}, 2017, pp. 5018--5027.

\bibitem{yang2023tvt}
Jinyu Yang, Jingjing Liu, Ning Xu, and Junzhou Huang,
\newblock ``Tvt: Transferable vision transformer for unsupervised domain adaptation,''
\newblock in {\em Proceedings of the IEEE/CVF Winter Conference on Applications of Computer Vision}, 2023, pp. 520--530.

\bibitem{kingma2014adam}
Diederik~P Kingma and Jimmy Ba,
\newblock ``Adam: A method for stochastic optimization,''
\newblock {\em arXiv preprint arXiv:1412.6980}, 2014.

\bibitem{zheng2022prompt}
Zangwei Zheng, Xiangyu Yue, Kai Wang, and Yang You,
\newblock ``Prompt vision transformer for domain generalization,''
\newblock {\em arXiv preprint arXiv:2208.08914}, 2022.

\bibitem{gao2021clip}
Peng Gao, Shijie Geng, Renrui Zhang, Teli Ma, Rongyao Fang, Yongfeng Zhang, Hongsheng Li, and Yu~Qiao,
\newblock ``Clip-adapter: Better vision-language models with feature adapters,''
\newblock {\em arXiv preprint arXiv:2110.04544}, 2021.

\end{thebibliography}

\end{document}